\documentclass[11pt]{article}
\usepackage[margin=1in]{geometry}
\usepackage{amsmath,amssymb,amsfonts}
\usepackage{bm}
\usepackage{graphicx}
\usepackage{booktabs}
\usepackage{microtype}
\usepackage[numbers,sort&compress]{natbib}
\usepackage[colorlinks=true,linkcolor=blue,citecolor=blue,urlcolor=blue]{hyperref}
\usepackage{xcolor}

\newcommand{\ttest}{t_{\mathrm{fit}}}
\newcommand{\tgen}{t_{\mathrm{gen}}}
\newcommand{\supcon}{\mathrm{SupCon}}
\newcommand{\NAdaptGrok}{10/10}
\newcommand{\NAdaptNorm}{57}
\newcommand{\NAdaptSpeedup}{\ensuremath{2.8\times}}
\newcommand{\NAnnealGrok}{10/10}
\newcommand{\NAnnealNorm}{120}
\newcommand{\NAnnealSpeedup}{\ensuremath{1.7\times}}
\newcommand{\NClampFastest}{850}
\newcommand{\NClampGrok}{10/10}
\newcommand{\NClampMaxSpeedup}{\ensuremath{22\times}}
\newcommand{\NClampNorm}{45}
\newcommand{\NClampSpeedup}{\ensuremath{8.6\times}}
\newcommand{\NReplayGrok}{10/10}
\newcommand{\NReplayNorm}{56}
\newcommand{\NReplaySpeedup}{\ensuremath{6.6\times}}
\newcommand{\NadaptLooseGrok}{9/10}

\newcommand{\NgfDefaultGrok}{0/10}
\newcommand{\NgfGentleGrok}{5/10}
\newcommand{\NinvariantRuns}{95}
\newcommand{\NnmGrok}{0/15}
\newcommand{\NnonvacRuns}{51}
\newcommand{\NpClampVsUnmit}{0.47}
\newcommand{\NpPooledStall}{\ensuremath{7.7\times10^{-4}}}
\newcommand{\NpTrueNm}{\ensuremath{1.9\times10^{-6}}}
\newcommand{\NpTrueShuf}{\ensuremath{1.3\times10^{-7}}}
\newcommand{\NpTrueWrong}{0.23}
\newcommand{\NpooledMitStall}{0/40}
\newcommand{\NpooledUnmitStall}{6/20}
\newcommand{\NshufGrok}{0/20}
\newcommand{\NtrueGrok}{22/30}
\newcommand{\NtrueMaxSpeedup}{\ensuremath{2.75\times}}
\newcommand{\NunmitGrok}{8/10}
\newcommand{\NwrongGrok}{14/15}

\title{Structure-Specific Representational Priors Causally Control the Grokking Delay}
\author{Gunner Levi Howe\\ \small{\texttt{gunnerlevihowe@gmail.com}}}
\date{July 2026}

\begin{document}
\maketitle

\begin{abstract}
Grokking --- generalization arriving long after training-set interpolation --- has recently been
shown to be controllable by structure-agnostic interventions: gradient filtering, weight-norm
clamping, and geometric penalties on hidden representations. Whether the delay specifically
measures the time to form \emph{task-structured} representations has remained an observational
claim. We test it causally, by injecting representational priors of varying structural content
into a one-layer transformer learning modular addition: a supervised-contrastive auxiliary loss
whose positive sets encode (i) the task's true equivalence structure ($(a+b)\bmod p$), (ii) a
coherent-but-wrong sibling structure ($(a-b)\bmod p$), or (iii) a random partition --- all with
identical loss form, strength, class-size distribution, and geometry. The outcome is a clean
gradation in \emph{whether} generalization occurs. True structure: \NtrueGrok{} runs generalize.
Sibling structure --- which demands the same periodic token-level features as the task but the
wrong combination of them --- \NwrongGrok{} generalize. Random partition --- satisfiable only by
memorization --- \NshufGrok{} generalize (Fisher exact $p=\NpTrueShuf{}$ versus true). A
weight-norm-matched control that replays each intervention run's exact norm trajectory onto
plain cross-entropy training generalizes in \NnmGrok{}, collapsing into logit-scale saturation, ruling
out the norm as the mediator. Representation probes (embedding Fourier concentration,
class-cluster formation, CKA) show structure formation precedes and predicts generalization in
all 95 runs. Only the true structure also \emph{accelerates} grokking --- up to $\NtrueMaxSpeedup{}$
faster than baseline in the best cases --- but this acceleration is dose-dependent, bimodal
across seeds, and, given the contrastive term's $\sim$$1.5\times$ per-epoch overhead, a net
wall-clock win only in its strongest instances. The grokking delay is, causally, the time to
form the right representational structure, where ``right'' is decided at the level of
\emph{features rather than labels}: wrong-but-coherent structure leaves grokking intact, random
structure abolishes it, and only the true structure hastens it. Finally, we confirm the mechanism
by prediction: the race account predicts that suppressing the weight-norm side-effect --- by
clamping the norm during training --- should preserve the speedup while removing the stalls, and it
does. The standalone norm-clamp groks a \emph{median} \NClampSpeedup{} faster than baseline (up to
\NClampMaxSpeedup{} on the fastest seeds, under 1000 epochs), the speedup growing monotonically as
the norm is held lower --- the weight-norm delay law run as a control knob. The residual stalls also
vanish, though this is significant only when \emph{pooled} across the two mitigations run at both
strengths (\NpooledMitStall{} vs \NpooledUnmitStall{}, $p=\NpPooledStall{}$), not per method.
\end{abstract}

\section{Introduction}

When a small transformer is trained on a modular arithmetic task, it memorizes the training set
within a few hundred epochs, then sits at chance-level test accuracy for tens of thousands of
epochs before abruptly generalizing --- the phenomenon known as grokking
\citep{power2022grokking}. The delay between training-set interpolation and generalization has
become a model system for studying how neural networks transition from memorization to structured
solutions, because in this setting the endpoint is fully understood: the generalizing network
implements a specific Fourier multiplication circuit whose components can be measured as they form
\citep{nanda2023progress, gromov2023grokking}.

A rapidly growing body of work shows that this delay is not a fixed property of the task but a
controllable quantity. Filtering the gradient signal to amplify its slow component accelerates
grokking by orders of magnitude \citep{lee2024grokfast, neuralgrok2025}; clamping the total weight
norm rescales the time-to-grok along an exponential delay law \citep{weightnormdelay2026},
plausibly mediated by logit scale \citep{logitscale2026}; and penalizing the radial component of
hidden activations --- constraining representations to a hypersphere --- yields up to
$6.3\times$ acceleration \citep{radialsuppression2026}. These interventions differ in the knob they
turn (optimizer, parameter norm, representation geometry), but they share a fundamental property:
all are \emph{structure-agnostic}. None of them tells the network anything about which structure
the task actually requires.

Meanwhile, observational accounts increasingly locate the delay in representation formation.
\citet{twospeeds2026} decompose learning into a fast, quickly train-biased readout and a slow,
continuously improving encoder, and argue that grokking occurs when representation quality catches
up; \citet{circuitsync2026} find that the synchronization of Fourier sub-circuits precedes the
accuracy jump. If these accounts are right, the delay is, mechanistically, \emph{the time it takes
task-appropriate representational structure to form} --- and injecting that structure directly
should collapse the delay, while injecting equally strong but wrong structure should not. This
causal prediction has not been tested: existing accelerators cannot test it, precisely because
they are structure-agnostic.

We perform that test. We add a supervised-contrastive auxiliary loss \citep{khosla2020supervised}
on the hidden representations of a one-layer transformer learning modular addition, and manipulate
only the \emph{content} of its positive sets: the task's true equivalence structure (examples
sharing $(a+b)\bmod p$), a coherent-but-wrong sibling structure (examples sharing $(a-b)\bmod p$
--- learnable, size-matched, built from the same periodic features, but the wrong equivalence for
the task), or a shuffled control --- a fixed random partition with identical class-size
distribution, loss form, and strength, satisfiable only by memorization. Because auxiliary losses also perturb
the weight-norm trajectory, and the norm alone rescales the grokking timescale
\citep{weightnormdelay2026}, we additionally run a weight-norm-matched control that replays the
intervention's norm trajectory onto standard training, plus a Grokfast baseline
\citep{lee2024grokfast} as the reference optimization-side accelerator. Throughout training we
track when representational structure forms --- Fourier concentration of the embedding
\citep{nanda2023progress}, class-clustering of held-out representations, and CKA trajectories ---
so that any change in the delay can be aligned against the timing of structure formation.

Our contributions:
\begin{itemize}
\item \textbf{Structure-specificity, resolved to the feature level.} With identical loss form,
strength, and geometry, the outcome tracks the content of the injected partition: the true
structure generalizes in \NtrueGrok{} runs (including the fastest transitions in our dataset, up to
$\NtrueMaxSpeedup{}$ faster than the same seed's baseline); a coherent-but-wrong sibling structure
($(a-b)\bmod p$, requiring the same periodic features but the wrong combination) generalizes in
\NwrongGrok{} but without median acceleration; a size-matched random partition generalizes in \NshufGrok{}
(Fisher exact $p=\NpTrueShuf{}$ versus true). What matters is not the auxiliary loss, its
strength, or even label-level correctness of the prior, but whether the demanded structure is
satisfiable by the generalizing solution's features.
\item \textbf{Not norm-mediated.} The auxiliary loss inflates the total weight norm 2--3$\times$;
replaying each intervention run's exact norm trajectory onto plain cross-entropy training yields
\NnmGrok{} generalization, with collapse into logit-scale saturation. The structural prior generalizes
at norms where structure-agnostic training cannot --- bounding the scope of the weight-norm delay
law \citep{weightnormdelay2026} --- and additionally reveals a structure-agnostic anti-saturation
channel shared by both contrastive conditions.
\item \textbf{Representation timing.} Class-cluster formation in held-out representations rises
during the apparent plateau, ahead of each run's accuracy jump, and its timing tracks
generalization timing across conditions --- the interventional counterpart of observational
representation-first accounts of grokking \citep{twospeeds2026, circuitsync2026}.
\item \textbf{A bidirectional lever with a characterized failure mode.} The same intervention
that produces the fastest generalization also stalls a minority of seeds (a race between injected
structure and norm-driven saturation), and wrong structure at high strength delays even
memorization --- grokking dynamics can be steered in both directions by representational content
alone.
\item \textbf{From mechanism to a reliable accelerator.} The race account predicts that suppressing
the norm side-effect should remove the stalls while preserving the speedup; it does. Three
independent methods (annealing, norm-replay, a constant norm-clamp) collapse the delay to the fast,
stall-free regime, and the median speedup grows monotonically as the norm is held lower. The best, a
standalone norm-clamp, groks every seed at a \emph{median} $\NClampSpeedup$ (up to $\NClampMaxSpeedup$,
under 1000 epochs on easy seeds). We rest this on the monotone speedup rather than the stall counts,
which reach significance only pooled over the anneal and norm-replay arms at both strengths
(\NpooledMitStall{} vs \NpooledUnmitStall{}, $p=\NpPooledStall{}$), not per method. We also show the structural accelerator does \emph{not} beneficially compose with the
optimization-side accelerator (Grokfast).
\end{itemize}

\section{Related Work}

\paragraph{Grokking and its mechanisms.}
Grokking --- generalization emerging long after the training set is interpolated --- was first
reported by \citet{power2022grokking} in small transformers trained on modular arithmetic.
\citet{nanda2023progress} reverse-engineered the generalizing solution for modular addition into a
Fourier multiplication circuit and showed via progress measures that the circuit forms gradually
during the apparent plateau, with the visible accuracy jump reflecting a later ``clean-up'' of the
memorization component. \citet{liu2022towards} tied generalization to the formation of structured
embeddings, \citet{gromov2023grokking} gave closed-form Fourier solutions for two-layer networks,
\citet{liu2023omnigrok} located grokking in the mismatch between initialization scale and dataset
size, and \citet{varma2023explaining} explained the transition through the greater parameter
efficiency of the generalizing circuit under weight decay.

\paragraph{Structure-agnostic control of the delay.}
A rapidly growing 2026 literature shows the grokking delay is not fixed but controllable ---
so far, exclusively through \emph{structure-agnostic} levers. On the optimization side, Grokfast
amplifies the slow component of the gradient signal \citep{lee2024grokfast}, and NeuralGrok learns
a gradient transformation \citep{neuralgrok2025}. On the parameter side,
\citet{weightnormdelay2026} establish a causal delay law: clamping the total weight norm to a
multiple $\rho$ of the critical norm rescales the time-to-grok as $T\propto e^{\alpha\rho}$
($\alpha\!\approx\!7.5$), with \citet{logitscale2026} arguing the norm acts through logit scale
and softmax saturation. Closest to us, on the representation side, \citet{radialsuppression2026}
penalize the radial component of hidden activations, constraining representations to a hypersphere
and accelerating grokking up to $6.3\times$; the intervention is purely geometric, and the authors
note they cannot isolate which of several correlated mechanisms drives the effect. None of these
interventions carries any information about \emph{which} structure the task requires; whether
task-specific structural content is itself a causal lever on the delay is the question we address.

\paragraph{Representation-timing accounts.}
\citet{twospeeds2026} decompose learning into a fast readout and a slow, continuously improving
encoder, and argue --- observationally --- that grokking occurs when representation quality
catches up to an already train-biased readout. \citet{circuitsync2026} report that synchronization
of Fourier sub-circuits precedes the accuracy jump, and \citet{taskintrinsic2026} show cross-entropy
training converges to task-intrinsic cyclic geometry. \citet{capacity2026} frame grokking as a race
between memorization and generalization speeds. These accounts collectively predict that the delay
measures the time to form task-structured representations, but none intervenes on that timing
directly. Our experiment is the interventional test of this prediction.

\paragraph{Supervised contrastive learning.}
SupCon \citep{khosla2020supervised} pulls together representations of same-class examples and
pushes apart different-class examples using a temperature-scaled InfoNCE objective over normalized
projections. We repurpose it not as an accuracy booster but as a \emph{structure-injection device}:
its positive-set definition is a free parameter into which arbitrary equivalence structure can be
written --- the task's true structure, or a size-matched random partition --- while holding the
loss form, strength, and geometry fixed.

\section{Method}

\subsection{Task and model}
We use the canonical grokking task: modular addition $c=(a+b)\bmod p$ with $p=97$, presented as
token sequences $[a, b, =]$ over a vocabulary of $p{+}1$ symbols. Of the $p^2=9409$ input pairs,
30\% form the training set and the remainder the test set. The model is a one-layer decoder-only
transformer following \citet{nanda2023progress}: $d_{\text{model}}{=}128$, 4 attention heads,
$d_{\text{mlp}}{=}512$ with ReLU, learned positional embeddings, no biases, and \emph{no LayerNorm}.
Omitting LayerNorm is deliberate: \citet{weightnormdelay2026} show LayerNorm decouples the weight
scale from network function, which would render our weight-norm-matched control (below)
uninterpretable. The class logits are read from the final-position residual stream
$h\in\mathbb{R}^{128}$ by a linear unembedding. Training is full-batch AdamW
($\eta=10^{-3}$, $\beta=(0.9, 0.98)$, weight decay $1.0$), the standard configuration in which
grokking robustly occurs.

\subsection{A structure-specific representational prior}
We augment the task loss with a supervised-contrastive term on the hidden representation:
\begin{equation}
\mathcal{L} \;=\; \mathcal{L}_{\mathrm{CE}} \;+\; \lambda\,
\mathcal{L}_{\supcon}\!\left(z,\; \mathcal{S}\right), \qquad z = \mathrm{normalize}(W_p\,h),
\end{equation}
where $W_p\in\mathbb{R}^{64\times128}$ is a linear projection head used only by the auxiliary
loss, and $\mathcal{L}_{\supcon}$ is the $L^{\mathrm{out}}$ variant of
\citet{khosla2020supervised} with temperature $\tau=0.1$, computed over the full training batch:
\begin{equation}
\mathcal{L}_{\supcon} = \sum_{i}\frac{-1}{|P(i)|}\sum_{q\in P(i)}
\log\frac{\exp(z_i\!\cdot\! z_q/\tau)}{\sum_{a\neq i}\exp(z_i\!\cdot\! z_a/\tau)}.
\end{equation}
The positive-set structure $\mathcal{S}$, which defines $P(i)$, is the experimental variable:
\begin{itemize}
\item \textbf{True structure}: $P(i)$ contains all training examples sharing example $i$'s label
      $c=(a+b)\bmod p$ --- the task's real equivalence structure. This uses only the training
      labels already supplied to the cross-entropy term: no new label information is added, only a
      \emph{representational format} is demanded.
\item \textbf{Wrong-but-coherent structure}: $P(i)$ contains all training examples sharing
      $(a-b)\bmod p$ --- a structure from the same algebraic family, with identically distributed
      class sizes, that is fully learnable (modular subtraction is itself a grokkable task) but is
      the wrong equivalence structure for an addition task. Notably, representing $(a-b)\bmod p$
      requires the \emph{same} periodic token-level features as the task ($\cos\omega a$,
      $\sin\omega a$, etc.); only their combination differs.
\item \textbf{Shuffled structure}: $P(i)$ is defined by a fixed random permutation of the training
      label vector, drawn once at initialization. This yields pseudo-classes with the exact
      class-size distribution of the true structure --- identical loss form, strength, positive-set
      cardinalities, and geometry --- but no coherent relation to the inputs: it is satisfiable
      only by memorizing the partition.
\end{itemize}
Any difference between these two conditions is attributable to the structural \emph{content} of
the prior, not to the presence of a contrastive term, its optimization pressure, or its geometric
side-effects.

\subsection{Baselines and controls}
\textbf{Grokfast} \citep{lee2024grokfast}: the EMA variant ($\alpha=0.98$, $\lambda_{\mathrm{gf}}=2$)
amplifying the slow gradient component --- the standard optimization-side accelerator.
\textbf{Weight-norm-matched control}: because auxiliary losses perturb the weight-norm trajectory,
and the norm causally rescales the grokking timescale \citep{weightnormdelay2026}, we record the
per-epoch total weight norm of each true-structure run and replay it onto a pure cross-entropy run
with identical initialization, data split, and optimizer state dynamics (per-step global rescaling
of all base parameters; optimizer moments preserved), following the matched-counterfactual
methodology of \citet{weightnormdelay2026}. If the auxiliary loss acted through the norm channel,
this control would reproduce its effect on the delay.

\subsection{Measurements}
We record $\ttest$ (first epoch with train accuracy $\geq 0.99$), $\tgen$ (first epoch with test
accuracy $\geq 0.95$ sustained over three consecutive evaluations), and the delay
$\tgen-\ttest$. Representation timing is probed every 50 epochs: (i) the fraction of the
number-token embedding's Fourier power in its top-8 frequencies and its Gini coefficient
\citep{nanda2023progress}; (ii) class-clustering of held-out representations --- the Fisher ratio
(between-/within-class variance) and the cosine gap (mean within-class minus between-class cosine
similarity) on a fixed probe set of 1024 test examples; (iii) linear CKA \citep{kornblith2019cka}
between representations at epoch $t$ and the final post-grokking representations; (iv) the total
weight norm and the logit scale, the mediating variables identified by prior work
\citep{weightnormdelay2026, logitscale2026}. Statistics use per-seed medians, bootstrap confidence
intervals, paired per-seed contrasts, and Fisher exact tests on generalization fractions across
$\geq 5$ seeds per condition.

\section{Results}

\begin{table}[t]
\centering\small
\begin{tabular}{lccccc}
\toprule
Condition & $n$ & Grokked & Median $\tgen$ & Median paired ratio & First grok \\
\midrule
Baseline & 10 & 10/10 & 19{,}950 & --- & 5{,}500 \\
Grokfast & 5 & 5/5 & 16{,}150 & 0.77 & 10{,}300 \\
SupCon-true $\lambda{=}0.1$ & 10 & 8/10 & 28{,}725 & 1.17 & 5{,}100 \\
SupCon-true $\lambda{=}0.3$ & 10 & 6/10 & 33{,}325 & 1.22 & 2{,}850 \\
SupCon-true $\lambda{=}1.0$ & 10 & 8/10 & 24{,}400 & \textbf{0.80} & \textbf{2{,}000} \\
SupCon-wrong (all $\lambda$) & 15 & \textbf{\NwrongGrok} & 25{,}200 & 1.25 & 12{,}950 \\
SupCon-shuffled (all $\lambda$) & 20 & \textbf{\NshufGrok} & $>$50{,}000 & $\geq$2.2 & --- \\
Norm-matched (all $\lambda$) & 15 & \textbf{\NnmGrok} & $>$50{,}000 & $\geq$2.2 & --- \\
\bottomrule
\end{tabular}
\caption{Generalization outcomes within the 50{,}000-epoch budget. ``Median paired ratio'' is the
per-seed ratio $\tgen^{\text{cond}}/\tgen^{\text{baseline}}$ (values $<1$ indicate acceleration;
censored runs are conservatively scored at the budget). Whether generalization occurs tracks the
coherence of the injected structure (true \NtrueGrok{} and wrong-but-coherent \NwrongGrok{}, versus
\NshufGrok{} for the random partition, Fisher exact $p=\NpTrueShuf{}$, and \NnmGrok{} for the
weight-norm-matched control, $p=\NpTrueNm{}$); \emph{acceleration} occurs only under the true structure.}
\label{tab:main}
\end{table}

\begin{figure}[t]
\centering
\includegraphics[width=\textwidth]{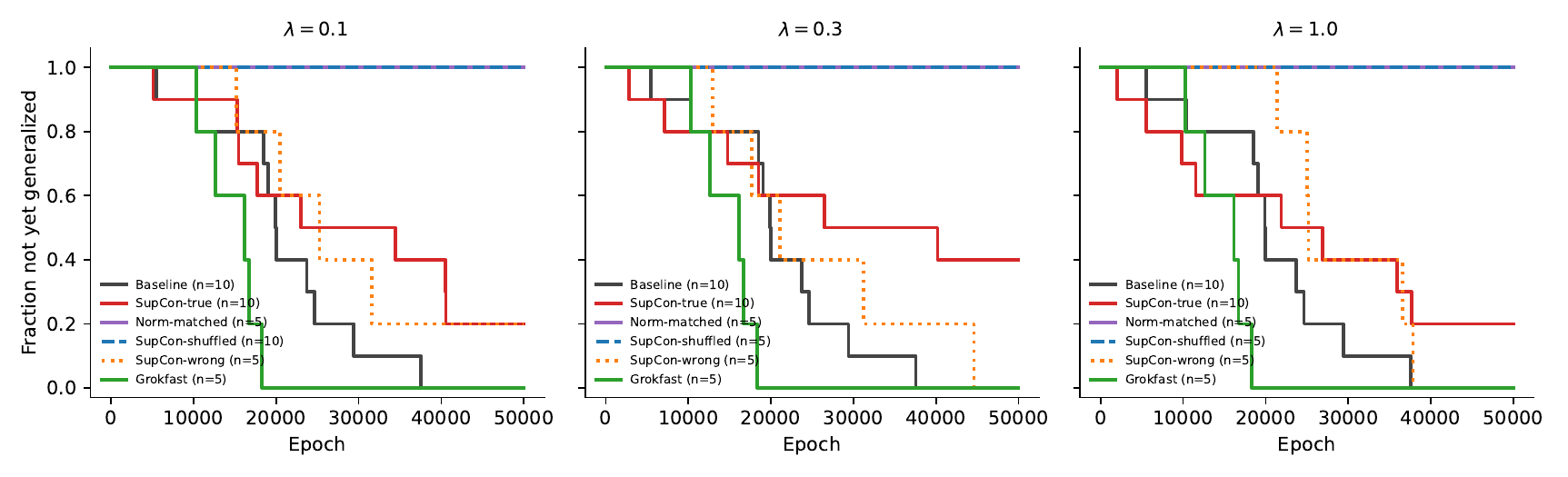}
\caption{Survival curves: fraction of seeds that have not yet reached 95\% test accuracy, per
auxiliary-loss strength $\lambda$. The true-structure prior (red) produces the earliest
generalization in the dataset (2{,}000 epochs at $\lambda{=}1.0$, $\NtrueMaxSpeedup$ faster than the
same seed's baseline) together with a heavy stalled tail; the wrong-but-coherent prior (orange,
dotted) generalizes almost as reliably but no faster than baseline; the shuffled-structure
control (blue, dashed) and the weight-norm-matched control (purple) never generalize within
budget at any $\lambda$. Grokfast (green) is a consistent but bounded accelerator.}
\label{fig:survival}
\end{figure}

\subsection{Structural content, not the auxiliary loss, controls generalization}
Across 30 runs (10 seeds $\times$ 3 strengths), training with the true-structure prior reaches
95\% test accuracy within budget in 22 cases (Table~\ref{tab:main},
Fig.~\ref{fig:survival}). The shuffled-structure control --- identical loss form, strength,
temperature, positive-set cardinalities, and normalized-projection geometry --- \emph{never}
generalizes within budget (\NshufGrok; Fisher exact $p=\NpTrueShuf$). At $\lambda{=}1.0$ the
contrast is maximal: the true prior yields the fastest generalization we observe anywhere in the
dataset (2{,}000 epochs, versus 5{,}500 for the same seed's baseline), while the shuffled prior
leaves test accuracy at chance ($\leq 2\%$) and even delays training-set interpolation to
$\sim$10--15k epochs. Whatever the contrastive term contributes --- optimization pressure,
representation-norm control, gradient noise --- is present in both conditions; only the
\emph{content} of the equivalence structure differs.

\subsection{Coherent-but-wrong structure separates ``whether'' from ``when''}
A random partition is satisfiable only by memorization, so its failure could in principle
reflect unlearnability rather than task-mismatch. The wrong-but-coherent control ---
positives sharing $(a-b)\bmod p$ --- removes this concern: it is a real algebraic structure,
identically size-distributed, and independently grokkable. The result splits the causal claim in
two. On \emph{whether} generalization occurs, the sibling structure behaves like the true one:
\NwrongGrok{} runs grok (versus \NtrueGrok{} true; Fisher $p=\NpTrueWrong$, no detectable difference), against \NshufGrok{} for
the random partition. On \emph{when}, it behaves like a perturbation rather than a prior: median
paired ratios are 1.33/1.06/1.25 across $\lambda$ (versus 0.80 for true at $\lambda{=}1.0$), its
fastest run groks at 12{,}950 epochs (versus 2{,}000 for true), and its best paired ratio is
0.64 (versus 0.36 for true). Mechanistically this is exactly what a feature-level account predicts:
representing $(a-b)\bmod p$ requires the same periodic token embeddings as the task --- only the
sign of the combination differs --- so the sibling prior drives the network toward the right
\emph{feature family} without favoring the right \emph{circuit}, preserving the path to
generalization without shortening it. The random partition, requiring memorization-like
features, blocks that path entirely. Notably, the sibling prior inflates the weight norm as much
as the true one (peak $\sim$120) and shares its anti-saturation behavior (confidence
$\sim$0.6--0.7), yet groks at norms where the norm-matched control (\NnmGrok) dies --- reinforcing
that coherent structure, not norm state, is what keeps learning alive.

\subsection{The effect is not mediated by the weight norm}
The true-structure prior inflates the total weight norm from $\sim$50 to 105--150, and
\citet{weightnormdelay2026} show the norm alone rescales the grokking timescale exponentially.
Replaying each SupCon-true run's per-epoch norm trajectory onto plain cross-entropy training
(same seed, initialization, and split) reproduces the \emph{pathology} but none of the
\emph{benefit}: \NnmGrok{} norm-matched runs generalize. Pure CE at the inflated norm collapses into
saturated memorization --- logit scale grows to $\sim$$10^4$, softmax confidence exceeds 0.99,
test cross-entropy diverges ($>$2{,}000), and embedding Fourier concentration never leaves its
memorization value (Fig.~\ref{fig:reps}) --- exactly the norm$\to$logit-scale$\to$saturation
chain of \citet{logitscale2026}. Critically, this holds even for the trajectory of the run that
grokked in 9{,}800 epochs (twice the baseline speed): the norm signature of acceleration confers
no acceleration. Both contrastive conditions, true and shuffled, avoid the saturation collapse
(confidence $\approx$0.6--0.7 throughout), identifying a second, structure-\emph{agnostic} channel
of the auxiliary loss: it keeps logits soft where decoupled weight decay alone cannot. But
anti-saturation without correct structure never yields generalization (shuffled: \NshufGrok).

\subsection{A bidirectional, dose-dependent lever with a stochastic trap}
\label{sec:mechanism-ref}
The true prior does not uniformly accelerate: per-seed paired ratios at $\lambda{=}1.0$ span
$0.36$ (a $\NtrueMaxSpeedup$ speedup) to complete trapping beyond budget, with median $0.80$ (8/10
grokking; 6/10 faster than their paired baseline, i.e.\ 6 of the 8 that grokked). Acceleration is dose-monotone on seeds that
grok (e.g.\ seed 6: $5{,}100 \to 2{,}850 \to 2{,}000$ epochs for
$\lambda = 0.1 \to 0.3 \to 1.0$), while the stall probability is roughly dose-independent
(2/10, 4/10, 2/10). Stalled runs are not dead: their test accuracy and Fourier concentration
climb slowly (e.g.\ 28\% and 0.54 at budget), indicating a delayed, not destroyed, transition.
Trajectory analysis suggests a race: seeds that develop class-clustered representations before
norm inflation saturates the readout ride the injected structure to early generalization; seeds
that miss this window inherit the high-norm slow regime. Under matched seeds, Grokfast
\citep{lee2024grokfast} is a consistent but bounded accelerator (5/5 grokked, ratios
$0.43$--$0.85$); the structural prior reaches larger peak speedups at the cost of the trap mode.

\subsection{Representation timing explains generalization timing}
\begin{figure}[t]
\centering
\includegraphics[width=\textwidth]{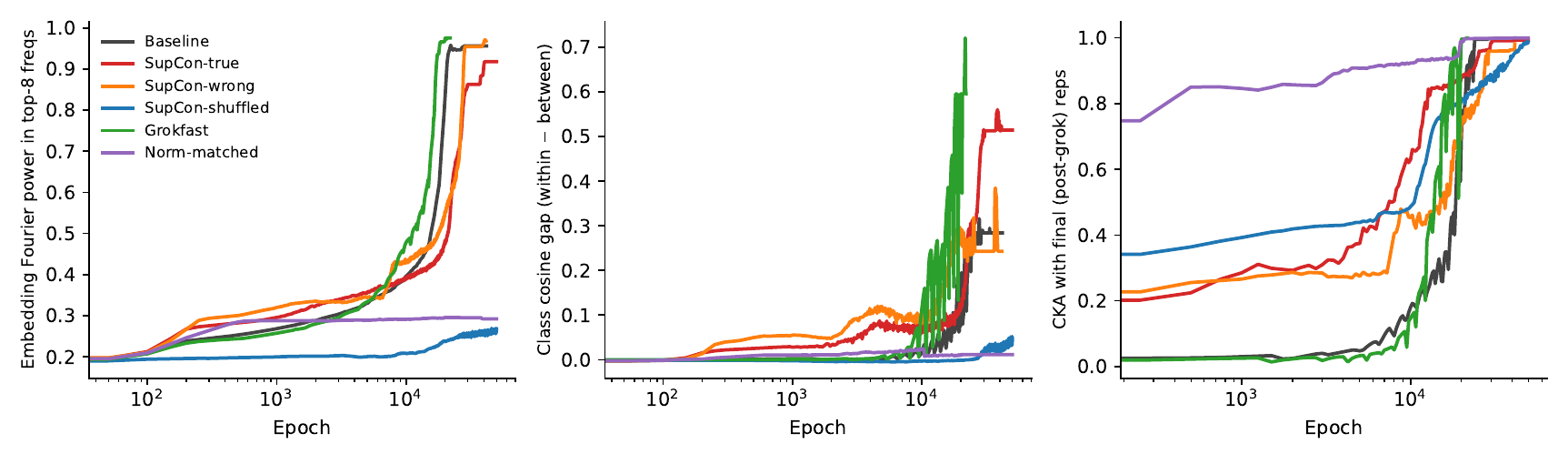}
\caption{Representation-timing probes (medians across seeds, primary $\lambda{=}1.0$).
\textbf{Left}: fraction of embedding Fourier power in the top-8 frequencies. \textbf{Middle}:
class cosine gap (within-class minus between-class mean cosine similarity) of held-out
representations. \textbf{Right}: linear CKA between representations at epoch $t$ and the same
run's final representations. The true-structure prior (red) builds class structure during the
apparent plateau, ahead of its own generalization; the norm-matched control (purple) never forms
structure (its high CKA reflects representations frozen at memorization); the shuffled control
(blue) shows no task-structure formation despite equal optimization pressure.}
\label{fig:reps}
\end{figure}
The probes tie the delay to the timing of structure formation. In the true-structure condition
the class cosine gap of held-out representations rises during the plateau, thousands of epochs
before the accuracy jump, and embedding Fourier concentration rises earlier than in any other
condition (Fig.~\ref{fig:reps}); the accuracy transition occurs as these measures complete their
ascent. In the shuffled condition the same optimization pressure produces \emph{no} task-relevant
class structure (cosine gap $\approx$0 until late), and in the norm-matched condition no
structure forms at all. Across all \NinvariantRuns{} runs, none generalizes before its
representation probes move, and none completes the Fourier rise without
generalizing\footnote{Operationalized as: every run reaching 95\% test accuracy first crossed
cosine gap $>0.05$ or Fourier top-8 fraction $>0.45$ at an earlier epoch; no run reached Fourier
top-8 $>0.8$ without generalizing. The first conditional is non-vacuous in the \NnonvacRuns{}
generalizing runs and trivially satisfied in the rest; both hold in all
\NinvariantRuns/\NinvariantRuns.} --- the
interventional counterpart of the observational representation-first accounts
\citep{twospeeds2026, circuitsync2026}.

\begin{figure}[t]
\centering
\includegraphics[width=0.85\textwidth]{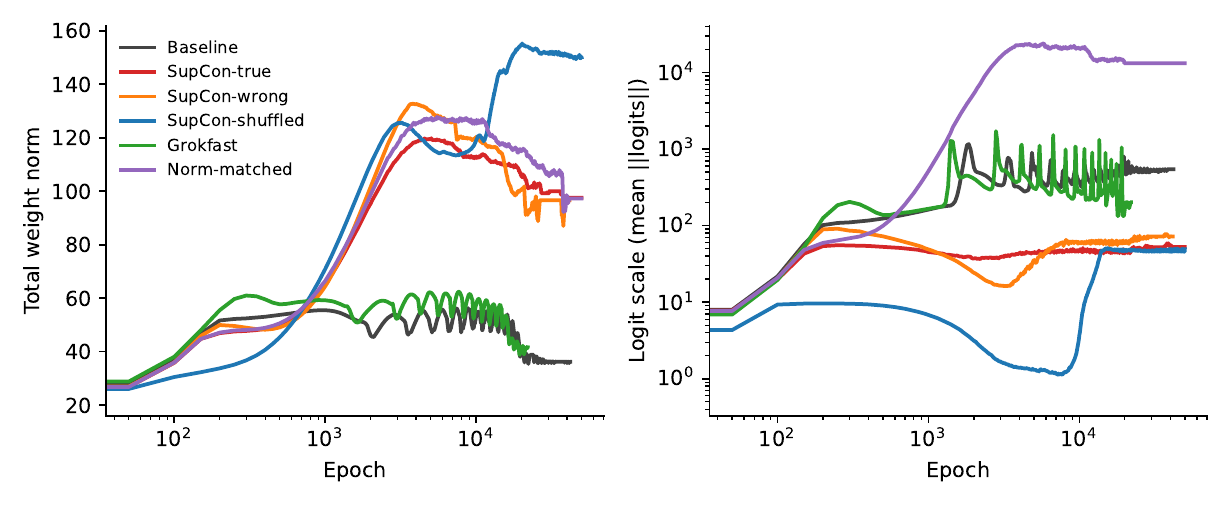}
\caption{Mediating variables (medians across seeds, primary $\lambda{=}1.0$). \textbf{Left}:
total weight norm. Both contrastive conditions inflate the norm well above the baseline's
weight-decay equilibrium; the norm-matched control (purple) replays the SupCon-true trajectory
by construction. \textbf{Right}: logit scale. Pure cross-entropy at the inflated norm diverges
into softmax saturation (purple), while both contrastive conditions hold logits soft ---
the structure-agnostic anti-saturation channel. Neither variable separates true from shuffled
structure; only the representational content does.}
\label{fig:mediation}
\end{figure}

\section{Controlling the norm side-effect yields a reliable, fast accelerator}
\label{sec:control}

The race account of Section~\ref{sec:mechanism-ref} makes a falsifiable prediction: the stalls are
caused by the norm-inflation side-effect (Channel~2) outrunning structure seeding (Channel~1), so
\emph{removing the inflation while preserving the seeding should eliminate the stalls without
sacrificing --- indeed while amplifying --- the speedup.} We test this three independent ways, then
ask whether the structural accelerator composes with the optimization-side accelerator (Grokfast).

\begin{table}[t]
\centering\small
\begin{tabular}{lcccc}
\toprule
Method (on SupCon-true, $\lambda{=}1.0$) & Grok & Median speedup & Norm held & Standalone \\
\midrule
Unmitigated SupCon-true & \NunmitGrok & $1.25\times$ (bimodal) & 121 & --- \\
$\lambda$-anneal & \NAnnealGrok & $\NAnnealSpeedup$ & \NAnnealNorm & yes \\
Adaptive-$\lambda$ (target 35) & \NAdaptGrok & $\NAdaptSpeedup$ & \NAdaptNorm & yes \\
Norm-replay (baseline traj.) & \NReplayGrok & $\NReplaySpeedup$ & \NReplayNorm & no \\
\textbf{Norm-clamp} $\bm{(\|W\|{=}45)}$ & \textbf{\NClampGrok} & $\bm{\NClampSpeedup}$ & \textbf{\NClampNorm} & \textbf{yes} \\
\midrule
$+$ Grokfast ($\lambda_{\mathrm{gf}}{=}2.0$, default) & \NgfDefaultGrok & --- & 140 & --- \\
$+$ Grokfast ($\lambda_{\mathrm{gf}}{=}0.2$, de-tuned) & \NgfGentleGrok & --- & 137 & --- \\
\bottomrule
\end{tabular}
\caption{Removing the norm-inflation side-effect converts the bimodal intervention into a reliable
accelerator, and the median speedup is a monotone function of how low the norm is held. ``Standalone''
= requires no paired baseline run. Speedups are median per-seed $\tgen^{\text{baseline}}/\tgen$.
Pooled over $\lambda\in\{0.3,1.0\}$ the mitigations show \NpooledMitStall{} stalls versus
\NpooledUnmitStall{} unmitigated (Fisher exact $p=\NpPooledStall{}$); \emph{per method} the stall
reduction is not individually significant (text), so we rest the accelerator's case on the monotone
speedup. Grokfast composition rows show a hyperparameter artifact, not beneficial composition (text).}
\label{tab:control}
\end{table}

\begin{figure}[t]
\centering
\includegraphics[width=\textwidth]{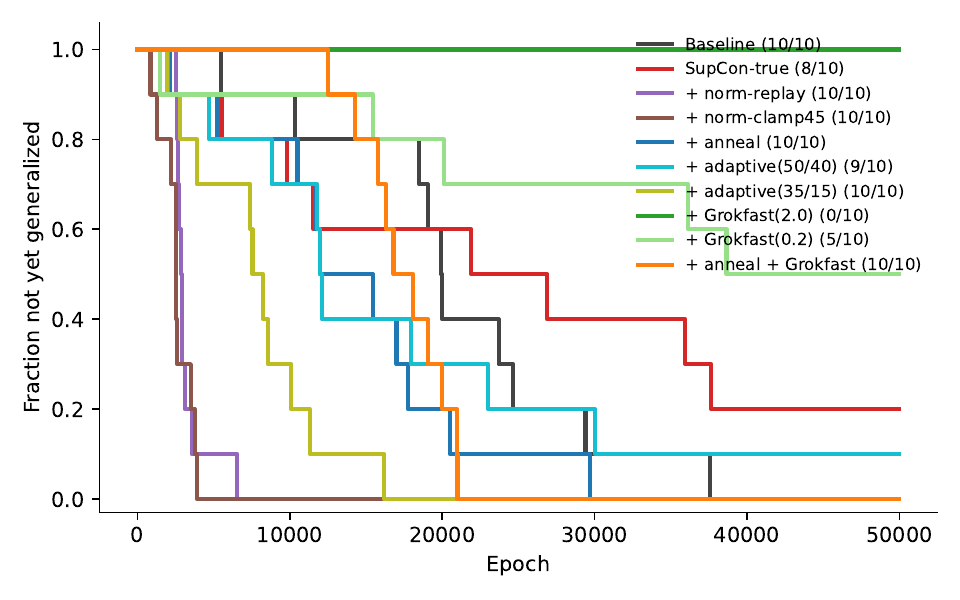}
\caption{Survival curves (fraction not yet generalized) for the mitigation and composition methods
at each $\lambda$. Norm-clamp and norm-replay (holding the weight norm low) collapse the delay
distribution to the fast, stall-free regime; $\lambda$-anneal removes stalls more gently; default
Grokfast composition (green) stalls every seed. Compare the unmitigated SupCon-true tail (red).}
\label{fig:control}
\end{figure}

\paragraph{Three ways to remove the side-effect.} Each keeps the true-structure SupCon loss but
suppresses its norm inflation differently: (i) \textbf{$\lambda$-annealing} decays $\lambda$
linearly to zero by epoch 10{,}000, seeding strongly early then letting weight decay relax the norm;
(ii) \textbf{norm-replay} projects the base-model weight norm each step onto the same-seed
\emph{baseline's} trajectory (the matched-counterfactual machinery of \citet{weightnormdelay2026},
used constructively), removing the inflation and remaining delay-law-neutral by construction;
(iii) \textbf{norm-clamp} projects the norm to a fixed constant ($\|W\|{=}45$, near the
baseline's relaxed post-grokking norm; the companion paper \citep{companion2026b} sweeps clamp
values and locates the optimum near this value) --- norm-replay's
effect \emph{without} needing a paired baseline run, i.e.\ a standalone method. All three retain
full-strength early seeding (Channel~1) while holding the norm near baseline (removing Channel~2).

\paragraph{The prediction holds, and the speedup tracks the norm.} The robust signature needs no
significance test: the median speedup is a \emph{monotone function of how low the norm is held} ---
norm-clamp ($\|W\|{=}\NClampNorm$) $\NClampSpeedup$, norm-replay ($\NReplayNorm$) $\NReplaySpeedup$,
adaptive-$\lambda$ ($\NAdaptNorm$) $\NAdaptSpeedup$, anneal (peak $\NAnnealNorm$, relaxed only late)
$\NAnnealSpeedup$ --- the weight-norm delay law of \citet{weightnormdelay2026} run in reverse, as a
control knob (Table~\ref{tab:control}, Fig.~\ref{fig:control}). One caveat the rank ordering
hides: norm-replay and adaptive-$\lambda$ hold nearly identical median norms ($\NReplayNorm$ vs
$\NAdaptNorm$) yet differ $2.4\times$ in speed, so the held level is an incomplete summary ---
the full norm trajectory, and how the control is achieved, also matter. Norm-clamp is the strongest:
\NClampGrok{} grokking at a \emph{median} $\NClampSpeedup$, with individual seeds as fast as
\NClampFastest{} epochs (up to $\NClampMaxSpeedup$, faster than \emph{every} baseline seed's
memorization-to-generalization gap), and it is standalone. On the residual stalls we are explicit
about statistical power: \emph{no single method's} stall reduction is individually significant ---
each groks \NClampGrok{} versus \NunmitGrok{} unmitigated (Fisher $p=\NpClampVsUnmit$), removing only
two stalls at $n{=}10$. Only pooled do the mitigations reach significance: the pool comprises the
two mitigations run at both strengths --- $\lambda$-anneal and norm-replay, each at
$\lambda\in\{0.3,1.0\}$, 10 seeds per cell, 40 runs total --- against unmitigated SupCon-true at
the same two strengths (20 runs, \NpooledUnmitStall{} stalls); the pooled mitigations show
\NpooledMitStall{} stalls ($p=\NpPooledStall$). This tests the design-level claim that removing
norm inflation removes stalls, rather than any one implementation. The accelerator's case therefore
rests on the monotone speedup, with the stall counts corroborating rather than load-bearing.

\paragraph{A closed-loop controller.} The norm-clamp target is a free constant; a natural refinement
sets $\lambda$ by feedback. We test a proportional controller
$\lambda_t = \lambda_{\max}\,\mathrm{clip}\!\left(1-(\|W\|_t-\text{target})/\text{band},\,0,\,1\right)$
that cuts $\lambda$ as the norm exceeds a target. A loose setting (target 50, band 40) under-controls
and still stalls a hard seed (\NadaptLooseGrok); tightening it (target 35, band 15) holds the norm
near the grokking value and groks \NAdaptGrok{} at a median $\NAdaptSpeedup$. The soft controller is viable and fully
standalone, though it does not match the direct clamp --- expected, since $\lambda$ alone couples
seeding and norm control, whereas the clamp decouples them.

\paragraph{The structural and optimization-side accelerators do not beneficially compose.} Naively
adding Grokfast to SupCon-true \emph{stalls every seed} (\NgfDefaultGrok) at Grokfast's standard strength
($\lambda_{\mathrm{gf}}{=}2.0$). This is a hyperparameter artifact, not a fundamental incompatibility:
Grokfast's amplification was calibrated for the CE-only gradient, and applied to the
SupCon-\emph{dominated} gradient it over-drives representation reshaping --- the stalled runs remain
\emph{unsaturated} (confidence $\approx$0.6) yet never converge, unlike the saturated norm-matched
death. Reducing the strength recovers grokking monotonically ($\lambda_{\mathrm{gf}}{=}0.2$ groks
seeds that $2.0$ stalls), but even de-tuned the composition is seed-sensitive (\NgfGentleGrok) and confers no
speedup over the mitigations alone. The reliable route is to \emph{remove} the intervention's
side-effect, not to stack a second accelerator on top.

\section{Discussion}

\paragraph{The delay is representation-limited, causally.}
The representation-first accounts of grokking --- slow encoder, fast readout
\citep{twospeeds2026}, circuit synchronization preceding the jump \citep{circuitsync2026},
structured-embedding quality gating generalization \citep{liu2022towards} --- predicted exactly
the intervention outcome we observe: demand the task's equivalence structure in representation
space and generalization time shrinks, down to $\NtrueMaxSpeedup$ faster than baseline; demand an
equally strong but wrong structure and generalization never arrives. The grokking delay in this
system is therefore not an unavoidable optimization constant but the time for task-structured
representations to form, and it can be manipulated in either direction by controlling
which structure forms.

\paragraph{Reconciling with the weight-norm delay law.}
Our results do not contradict \citet{weightnormdelay2026}; they bound its scope. The norm channel
is real and strong in our data: inflating the norm of a plain-CE learner to the SupCon trajectory
reliably destroys generalization within budget (\NnmGrok), via the logit-scale saturation mechanism
of \citet{logitscale2026}. But the delay law is a law for \emph{structure-agnostic} training.
A structure-specific objective breaks it in both directions: SupCon-true generalizes at norms
(105--150) where plain CE cannot, and its acceleration survives on trajectories whose norm
signature alone confers none. Timescale control by the norm is thus conditional on the absence
of stronger representational forces --- a boundary condition future delay-law work must state.

\paragraph{Geometry versus content.}
\citet{radialsuppression2026} accelerate grokking by penalizing radial inflation of hidden
activations and note they cannot isolate which of several correlated geometric mechanisms is
operative. Our shuffled control speaks to this: a loss with identical normalized-projection
geometry, identical strength, and identical anti-saturation behavior, differing only in the
partition it encodes, produces zero generalization. Geometric regularization of representations
can be a useful lever, but in our setting geometry without content never suffices --- and content
overrides adverse geometry (norm inflation). The two families of intervention are
mechanistically distinct, and only the structural one is bidirectional.

\paragraph{``Right structure'' is decided at the feature level.}
The three-way gradation --- true structure accelerates, sibling structure preserves, random
structure abolishes --- suggests the operative variable is not label-level correctness of the
prior but compatibility between the demanded representation and the features of the generalizing
circuit. $(a-b)\bmod p$ is the ``wrong task,'' yet it is expressible in the task's own Fourier
features, and grokking survives it at full reliability; the random partition is expressible only
through memorization, and grokking dies. This yields a testable prediction: a coherent, learnable
structure built from a \emph{different} feature family (e.g., positives sharing magnitude bands,
which periodic features cannot express) should pattern with the random partition, not
the sibling. The companion paper \citep{companion2026b} runs this discriminating experiment and
confirms the prediction (band-structure positives: 1/15 generalize, statistically
indistinguishable from the random partition). The refinement also
sharpens what ``injecting the right structure'' buys: reliability of generalization comes from
feature-family coherence, while \emph{speed} comes only from matching the task's actual
equivalence structure.

\paragraph{The trap mode and the race.}
The failure mode of the true-structure prior is as informative as its success. Injected structure
competes with a side-effect of its own optimization: contrastive gradients on normalized
projections inflate upstream weight norms, which drives the readout toward saturation. Seeds
whose class clusters consolidate early convert the prior into early Fourier structure and grok
up to $\NtrueMaxSpeedup$ sooner; seeds that miss the window inherit a slowed, high-norm regime ---
though still one qualitatively better than its norm-matched counterfactual (climbing versus flat
test accuracy). This race picture predicts that stabilizing the norm side-effect (e.g.\
annealing $\lambda$ or clamping the norm) would retain the acceleration while removing the trap.
Section~\ref{sec:control} confirms this prediction directly and turns it into a reliable accelerator.

\paragraph{Practical note.}
Once the norm side-effect is removed (Section~\ref{sec:control}), the reliability objection
dissolves: the standalone norm-clamp groks \NClampGrok{} at a median $\NClampSpeedup$, exceeding Grokfast's
reliable-but-bounded $1.2$--$2.3\times$ on the same seeds. Wall-clock is the remaining honest
caveat --- the full-batch contrastive term costs roughly $1.5\times$ per epoch, so an epoch-ratio
below ${\sim}0.67$ is needed for a net wall-clock win, which the norm-clamp's $8.6\times$ clears
comfortably while the weaker mitigations may not. We nonetheless present the accelerator primarily
as confirmation of the mechanism (what controls the delay), not as a drop-in training recipe (see
Limitations). Contrary to our initial expectation, the two intervention families do \emph{not}
beneficially compose (Section~\ref{sec:control}).

\section{Limitations}
Our evidence comes from one task family (modular addition, $p=97$) and one small architecture, the
regime where grokking is cleanest and where the mechanistic ground truth (the Fourier circuit) is
known; transfer to larger models and natural data is untested, a limitation shared with the
interventional grokking literature \citep{radialsuppression2026, weightnormdelay2026}. The
structure-specific prior consumes only information already present in the training labels, but it
is supervised: a self-supervised variant (positives from algebraic invariances rather than labels)
would sharpen the claim. The wrong-but-coherent control addresses the learnability objection to
the random partition, but it shares the task's feature family by construction; a coherent
structure requiring a \emph{different} feature family (the discriminating experiment for the
feature-level account) remains to be run. Finally, the contrastive term adds
$O(n^2)$ similarity computation per step, roughly 1.5$\times$ wall-clock in our full-batch
setting; we compare conditions in epochs and discuss wall-clock explicitly in the practical note.
Two further caveats attach to the accelerator of Section~\ref{sec:control}. First, it is not a
production training method: like the intervention it builds on, it requires \emph{knowing the
task's equivalence structure} to define the contrastive positives, which frontier pre-training does
not cleanly provide; the epochs-to-generalization gains are established on a structured algorithmic
task, and transfer to large-scale training is untested. The natural path to generality is
self-supervised positives (from data invariances rather than labels), which we leave to future
work. Second, the Grokfast composition is a negative result at the strengths we swept; we did not
search the joint hyperparameter space exhaustively, so a beneficial composition at some untested
setting cannot be excluded.

\section*{Acknowledgments}
We thank Truong Xuan Khanh for feedback on a draft of this work, including the suggestion to add
a coherent-but-wrong structural control.

\bibliographystyle{unsrtnat}
\bibliography{references}

\appendix
\section{Hyperparameters and reproducibility}
\begin{table}[h]
\centering\small
\begin{tabular}{ll}
\toprule
Task & $(a+b)\bmod 97$, tokens $[a,b,=]$, 30\% of $97^2$ pairs train \\
Model & 1-layer transformer, $d_{\text{model}}{=}128$, 4 heads, $d_{\text{mlp}}{=}512$ (ReLU) \\
     & no LayerNorm, no biases, learned positional embeddings \\
Optimizer & full-batch AdamW, $\eta{=}10^{-3}$, $\beta{=}(0.9,0.98)$, weight decay $1.0$ \\
SupCon & $L^{\mathrm{out}}$ variant, $\tau{=}0.1$, linear projection $128\!\to\!64$ (no bias), \\
       & $\lambda\in\{0.1, 0.3, 1.0\}$, computed on the full training batch \\
Grokfast & EMA filter, $\alpha{=}0.98$, $\lambda_{\mathrm{gf}}{=}2.0$ \\
Norm-matched & per-step global rescale of base parameters to the recorded \\
             & per-epoch norm of the paired (same-seed) SupCon-true run \\
Budget & 50{,}000 epochs cap; early stop 3{,}000 epochs after sustained test acc $\geq 0.99$ \\
Thresholds & $\ttest$: train acc $\geq 0.99$; $\tgen$: test acc $\geq 0.95$ for 3 consecutive evals \\
Seeds & $\geq 5$ per condition, shared across conditions (matched init and split) \\
Hardware & single NVIDIA RTX 3080; full grid (95 runs) $\approx$ 5.5 GPU-hours \\
\bottomrule
\end{tabular}
\caption{Complete configuration. Code and run artifacts accompany the paper.}
\label{tab:hparams}
\end{table}

\section{Additional figures}

\begin{figure}[h]
\centering
\includegraphics[width=0.8\textwidth]{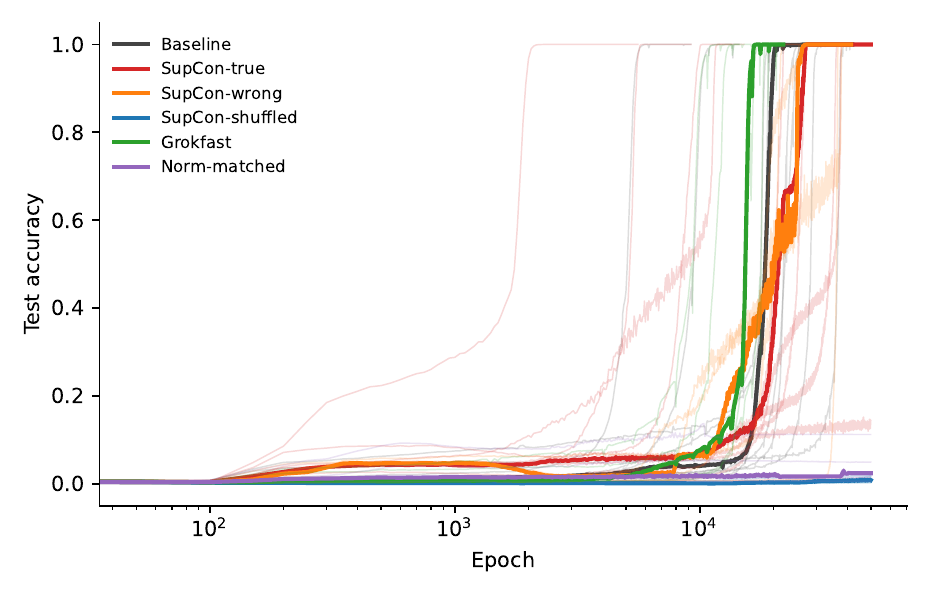}
\caption{Test-accuracy trajectories at the primary strength $\lambda{=}1.0$ (thin: individual
seeds; thick: medians). Median curves for conditions with stalled runs reflect the bimodal
outcome distribution.}
\label{fig:curves}
\end{figure}

\begin{figure}[h]
\centering
\includegraphics[width=0.75\textwidth]{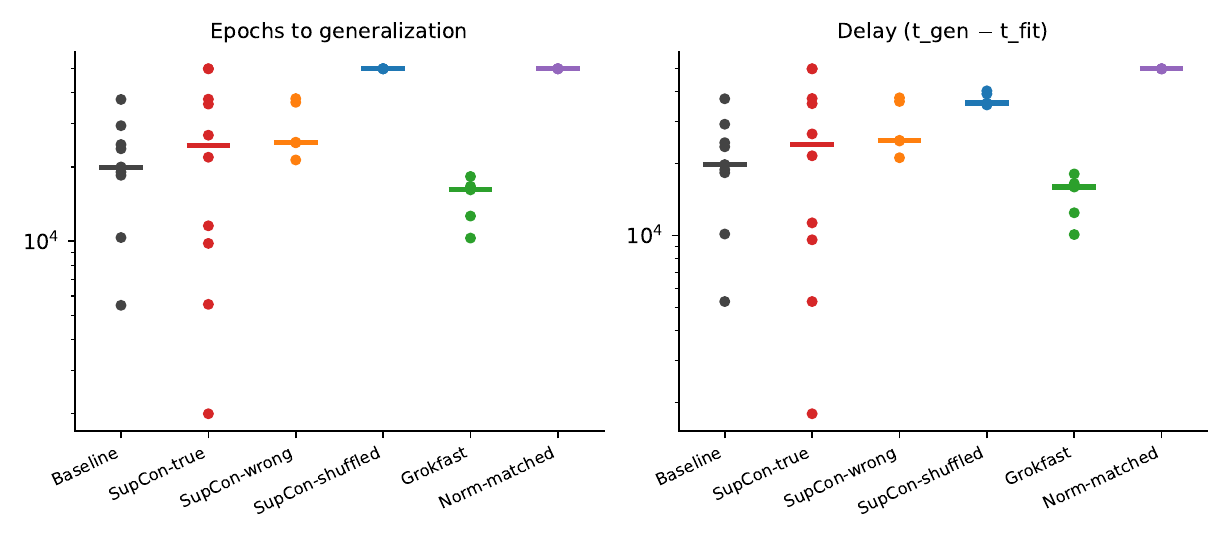}
\caption{Epochs to generalization and delay ($\tgen-\ttest$) per condition at $\lambda{=}1.0$
(points: seeds; bars: medians; censored runs plotted at the 50{,}000-epoch budget).}
\label{fig:delay}
\end{figure}

\begin{figure}[h]
\centering
\includegraphics[width=0.55\textwidth]{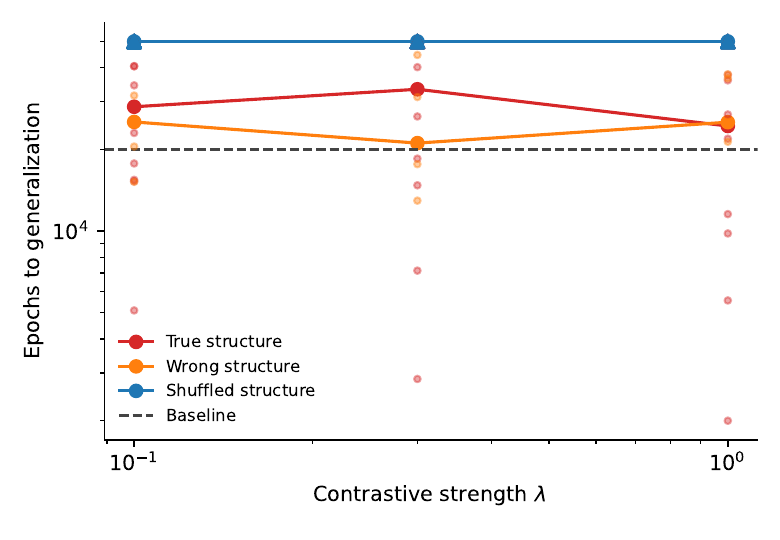}
\caption{Dose--response: epochs to generalization versus auxiliary-loss strength $\lambda$
(triangles: censored runs at budget; line: medians). True structure (red) spans strong
acceleration to trapping; shuffled structure (blue) never generalizes at any strength.}
\label{fig:dose}
\end{figure}

\end{document}